\documentclass[journal]{IEEEtran} 

\usepackage[english]{babel}

\usepackage{amsmath}
\usepackage{graphicx}
\usepackage[colorlinks=true, allcolors=blue]{hyperref}
\usepackage{amssymb}
\usepackage{pifont}
\newcommand{\cmark}{\ding{51}}%
\newcommand{\xmark}{\ding{55}}%

\usepackage{multirow}
\usepackage[dvipsnames]{xcolor}
\usepackage{threeparttable}

\usepackage{subcaption}
\usepackage{svg}

\usepackage{breqn}

\DeclareMathOperator*{\argmin}{\mbox{arg\,min}}

\usepackage{xcolor}

\title{NR-SLAM: Non-Rigid Monocular SLAM}
\author{Juan J. Gómez Rodríguez,
        José M.M. Montiel,~\IEEEmembership{Member,~IEEE} and Juan D. Tardós,~\IEEEmembership{Fellow,~IEEE}
\thanks{This work was supported by EU-H2020 grant 863146: ENDOMAPPER, Spanish government grant PID2021-127685NB-I00 and by Aragón government grant DGA\_T45-17R and PhD scholarship of J. J. Gómez-Rodrígez.}
\thanks{The authors are with the Instituto de Investigaci\'on en Ingenier\'ia de Arag\'on (I3A), Universidad de Zaragoza, 
Mar\'ia de Luna 1, 50018 Zaragoza, Spain. E-mail: \{jjgomez, josemari, tardos\}@unizar.es.}
}

\begin{document}
\maketitle

\begin{abstract}
In this paper we present NR-SLAM, a novel non-rigid monocular SLAM system founded on the combination of a Dynamic Deformation Graph with a Visco-Elastic deformation model. The former enables our system to represent the dynamics of the deforming environment as the camera explores, while the later allows us to model general deformations in a simple way.

The presented system is able to automatically initialize and extend a map modeled by a sparse point cloud in deforming environments, that is refined with a sliding-window Deformable Bundle Adjustment. This map serves as base for the estimation of the camera motion and deformation and enables us to represent arbitrary surface topologies, overcoming the limitations of previous methods.

To assess the performance of our system in challenging deforming scenarios, we evaluate it in several representative medical datasets. In our experiments, NR-SLAM outperforms previous deformable SLAM systems, achieving millimeter reconstruction accuracy and bringing automated medical intervention closer. For the benefit of the community, we make the source code public.
\end{abstract}

\section{Introduction}
Visual Simultaneous Localization and Mapping  (V-SLAM) techniques have been used in the last decade in a wide range of applications to locate an agent, from autonomous robots to augmented/virtual reality devices, in unknown environments. While these application may seem very different, they share a basic assumption that is crucial for V-SLAM techniques: the rigidity of the environment. While simple, this assumption allows to apply multi-view geometry to reconstruct the environment and locate the camera.

However, there are multiple applications in which the environment cannot be assumed to be stationary. Imagine a Minimal Invasive Surgery (MIS) procedure that aims to remove a polyp. By applying V-SLAM algorithms, the surgeon could be guided to the exact polyp location, easing the intervention. However, medical imagery presents scenarios in which the environment can freely deform due to the intervention, breathing or heartbeats. This simple change in the application domain renders current V-SLAM systems unable to produce accurate and faithful reconstructions. Moreover, most medical imagery is collected with a single monocular camera, which makes the problem of V-SLAM in deformable scenarios even harder, as no geometrical 3D information can be collected from the sensor. MIS imagery poses several additional issues, like poor texture and challenging illumination, worsening the data association step in any V-SLAM system. 

If that was not enough, deformable V-SLAM poses another challenge: how to represent in a compact and general way deforming surfaces. Some approaches use Signed Distance Functions (SDFs) implemented with 3D voxels, which do not scale well while exploring. Other approaches use a triangle mesh that scales better to bigger maps, but assumes planar topology, excluding tubes or surfaces with holes. Finally, most previous works assume isometric of quasi-isometric deformations, which is very questionable in many medical applications. All these challenges make monocular V-SLAM in medical sequences an open problem without a known general solution. 

\begin{figure}
\centering
  \includegraphics[width=\columnwidth]{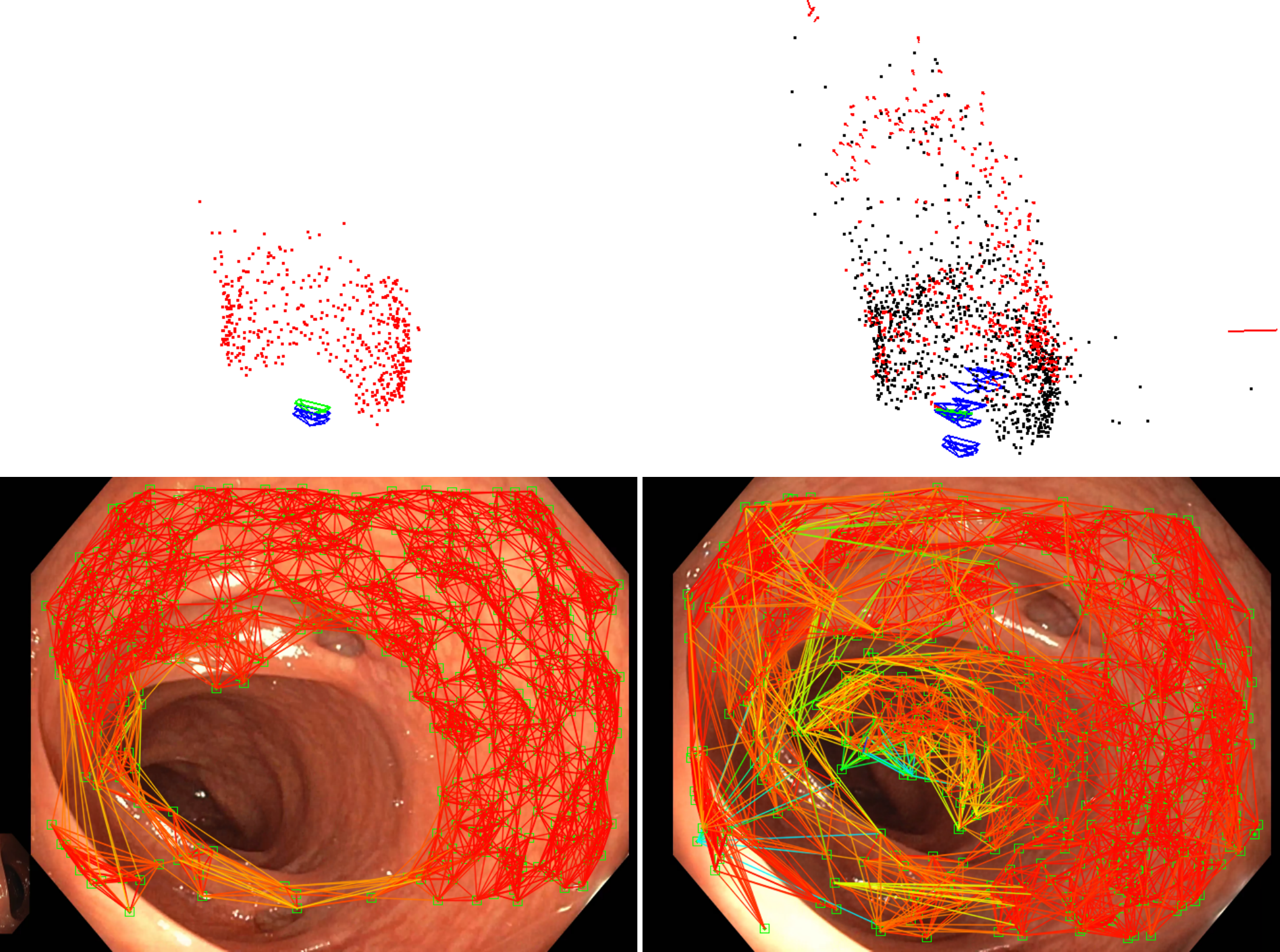}
  \caption{Reconstruction example of NR-SLAM in a real colonoscopy from the Endomapper dataset \cite{azagra2022endomapper}.}
  \label{fig::teaser}
\end{figure}

\begin{figure}
\centering
  \includegraphics[width=\columnwidth]{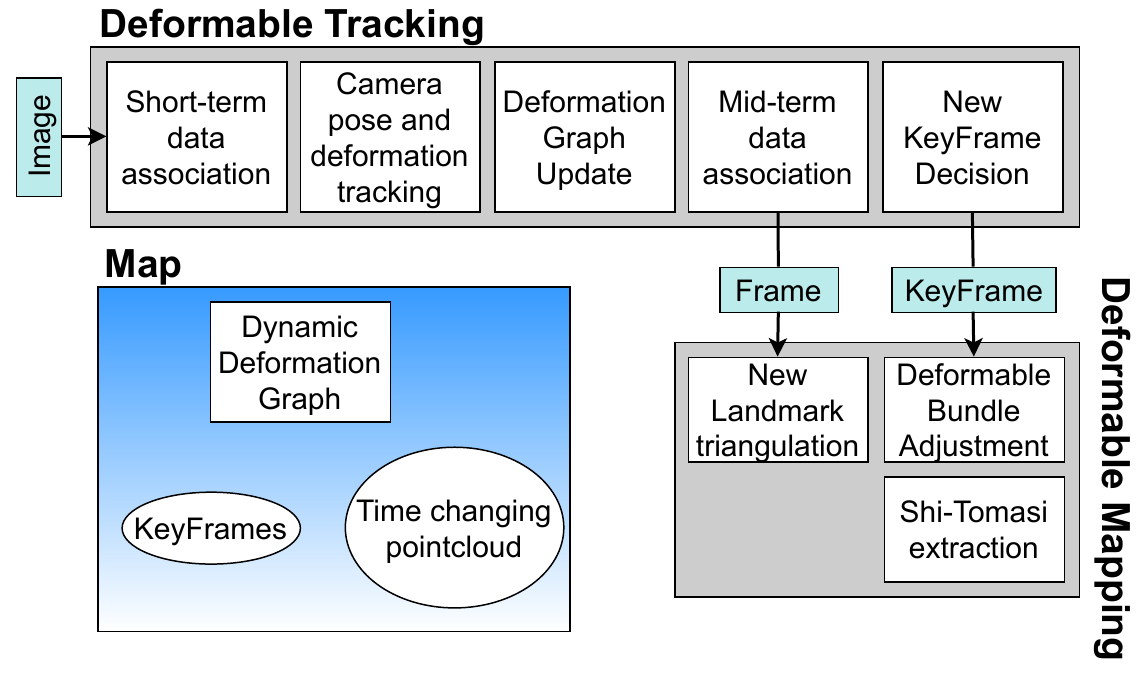}
  \caption{Main system components of NR-SLAM.}
  \label{fig::nrslam}
\end{figure}

In this work we propose NR-SLAM, a novel non-rigid monocular SLAM system that copes with the above limitations  and can be splitted in 3 main components (Fig. \ref{fig::nrslam}): deformable tracking, deformable mapping, and the map. The map represents the deformable environment with a time changing sparse point cloud that is easy to process and to scale when exploring new areas. Point maps are interconnected in a graph structure that relates points belonging to the same surface and that undergo similar deformations. On top of that, we model deformations as a set of pair-wise deformations with a simple mathematical representation. All these in conjunction with a robust and accurate data association based on optical flow make NR-SLAM able to reconstruct MIS imagery with unprecedented accuracy. To summarize, our contributions are:


\begin{itemize}
\item A complete monocular deformable SLAM system free from any topological and isometric assumptions.
\item A map composed of a sparse set of points and a novel Dynamic Deformation Graph (DDG) that relates map points that deform together (Fig. \ref{fig::teaser}), with a simple visco-elastic deformation model.
\item A robust and accurate semi-direct method for camera tracking and deformation estimation based on the DDG.
\item A deformable mapping method able to initialize, extend and refine the map estimations made by the deformable tracking.
\item Experimental validation in a relevant set of medical datasets. See the accompanying video for examples. 
\item For the benefit of the community, we release NR-SLAM as an open source library\footnote{\url{https://github.com/endomapper/NR-SLAM}}.
\end{itemize}


\section{Related work}
\begin{table*}[]
\centering
    \caption{Summary of the most representative methods for deformable SLAM, in chronological order.}\label{tab::def_systems}
\begin{tabular}{|l|c|c|c|c|c|c|}
\hline
                                                & Sensor     & Tracking? & Mapping? & Map Initialization     & Map Extension                                                             & Map representation                                                                       \\ \hline \hline
DynamicFusion \cite{newcombe2015dynamicfusion}  & RGB-D      & \cmark    & \cmark   & Depth                  & Depth                                                                     & Voxels                                                                                   \\ \hline
KillingFusion \cite{slavcheva2017killingfusion} & RGB-D      & \cmark    & \cmark   & Depth                  & Depth                                                                     & SDFs                                                                                     \\ \hline
VolumeDeform \cite{innmann2016volumedeform}     & RGB-D      & \cmark    & \cmark   & Depth                  & Depth                                                                     & SDFs                                                                                     \\ \hline
MIS-SLAM \cite{song2017dynamic}                 & Stereo     & \cmark    & \cmark   & Stereo                  & Stereo                                                                     & Sparse point cloud                                                                       \\ \hline
SurfelWarp \cite{gao2019surfelwarp}             & RGB-D      & \cmark    & \cmark   & Depth                  & Depth                                                                     & Surfels                                                                                  \\ \hline
\hline
DefSLAM \cite{lamarca2020defslam}                & Monocular  & \cmark    & \cmark   & Plane at unit distance & Isometric NRSfM                                                           & Triangle mesh                                                                            \\ \hline
SD-DefSLAM \cite{gomez2021sd}                   & Monocular  & \cmark    & \cmark   & Plane at unit distance & Isometric NRSfM                                                           & Triangle mesh                                                                            \\ \hline
RNN-SLAM \cite{ma2021rnnslam}             & \begin{tabular}[c]{@{}c@{}}Monocular + \\ depth network\end{tabular}      & \cmark    & \cmark   & Depth network                & Depth network                                                                      & Dense point cloud                                                                                  \\ \hline 
DSDT \cite{lamarca2021direct}                   & Monocular* & \cmark    & \xmark   & Stereo   & -                                                                         & Surfels                                                                                  \\ \hline
MCPD \cite{rodriguez2022tracking}               & Monocular  & \cmark    & \xmark   & Essential matrix       & -                                                                         & Sparse point cloud                                                                       \\ \hline
NR-SLAM (ours)                                         & Monocular  & \cmark    & \cmark   & Essential matrix       & \begin{tabular}[c]{@{}c@{}}Rigid/deformable \\ triangulation\end{tabular} & \begin{tabular}[c]{@{}c@{}}Sparse point cloud\\ + Dynamic Deformation Graph\end{tabular} \\ \hline
\end{tabular}
\end{table*}

In this section, we introduce a summary of the main V-SLAM algorithms for deformable environments (Table \ref{tab::def_systems}). We first introduce systems that use 3D cameras (stereo or RGB-D) and then pure monocular systems. 

\subsection{Non-Monocular deformable SLAM}
The use of 3D sensor is a common practice in literature to solve deformable SLAM, as the depth provided greatly reduces the complexity of the problem. DynamicFusion \cite{newcombe2015dynamicfusion} presented a deformable SLAM system with just a RGB-D camera. It builds a canonical model of the scene, i.e. its shape at rest, and deforms it to fit the current observed depth. For that, it combines a form of a Embedded Deformation graph model (ED) \cite{sumner2007embedded} with an as-rigid-as-possible regularizer. Fundamentally, ED models build a discretization of the deformations space in a graph structure, speeding up dense reconstructions. DynamicFusion inspired new works like VolumeDeform \cite{innmann2016volumedeform}, where the introduction of photometric image alignment improved the quality of the deformable reconstructions. Later, KillingFusion \cite{slavcheva2017killingfusion} proposed a new system in which deformations were required to be smooth and nearly isometric by using approximate Killing vector fields. MIS-SLAM \cite{song2017dynamic} presented a stereo deformable SLAM system based on as-rigid-as-possible deformations in conjunction with an ED model to solve for the deformations. However, the above methods present a common critical limitation: the map representation as Signed Distance Functions (SDFs). While useful, SDFs  scale poorly with the size of the map, rendering the methods not viable for exploration setups. This was addressed in Surfelwarp \cite{gao2019surfelwarp} where a surfel representation was used instead of the classical SDF.

\subsection{Monocular deformable SLAM}
Monocular deformable SLAM is a much more difficult problem, as Structure from Motion (SfM) with a set of freely moving points is unconstrained. The field has been dominated mainly by 2 group of methods: Shape-from-Template (SfT) and Non-Rigid Structure-from-Motion (NRSfM). The former aims to recover the deformed shape of an object imaged by a monocular camera with respect to a a shape-at-rest (template). The most effective approach in SfT is to assume isometric deformations as it has been proven to provide accurate real time solutions \cite{bartoli2015shape}, \cite{chhatkuli2016stable}.
The other family of methods, NRSfM, makes reconstructions of a deforming surface from a stream of monocular images without a known template. First solutions to this problem were formulated using orthographic cameras and statistical models in which the shape of the surface is a linear combination of low-dimensional basis models \cite{bregler2000recovering}, \cite{dai2014simple}. As the use of orthographic cameras is not feasible for real applications, later methods proposed to extend the isometric assumption with the use of perspective cameras \cite{chhatkuli2016inextensible}, \cite{parashar2017isometric} producing excellent results. The recent work presented in \cite{sengupta2022convex} relaxes the isometric constraint to an area-preserving (equiareal) deformation constraint. However, the isometric and equiareal assumptions are not applicable to a wide range of scenarios like living tissue in endoscopies. Moreover, SfT and NRSfM methods assume a static camera, not being directly applicable to deformable SLAM scenarios.

The first deformable monocular SLAM system, able to build and extend a deformable map, was DefSLAM \cite{lamarca2020defslam}. It was built on ORB-SLAM \cite{mur2015orb} by changing some parts of the pipeline to adapt it to the deformable case. First, a NRSfM algorithm \cite{parashar2017isometric} was integrated in the mapping thread to compute a surface template at keyframe rate. Secondly, those templates were aligned in the tracking thread and a SfT method \cite{lamarca2018camera} was used to jointly estimate the camera pose and deformations at frame rate. As these methods are very sensitive to spurious data, SD-DefSLAM \cite{gomez2021sd} proposed to solve the data association step in the SLAM pipeline using optical flow techniques, greatly improving the robustness and accuracy of the system, making it usable in medical sequences. However, these family of methods have two main limitations: the map representation as a triangle mesh and the isometric assumption. The triangle mesh makes these methods unable to reconstruct surfaces with discontinuities or holes. This was addressed in DSDT \cite{lamarca2021direct} by formulating the camera tracking and deformation estimation as a photometric alignment of isometric surfels. Nevertheless isometry does not model certain types of surfaces like living tissue and thus this method is not suitable for endoscopic use. In \cite{rodriguez2022tracking} we proposed to tackle both limitations by solving the camera tracking and deformation estimation problem by modelling surfaces as sparse point clouds restricted by as-rigid-as-possible deformations and an ED model. 

Although not a deformable SLAM system, it is worth mentioning the interesting work of RNN-SLAM \cite{ma2021rnnslam}, that combines DSO \cite{engel2017direct} with a recurrent neural network that estimates pose and depth from monocular images, obtaining the best reconstructions of colon sections so far. However, the method is specific for colonoscopies and ignores deformations. 

Following \cite{rodriguez2022tracking} and in contrast with previous approaches, we solve here the full monocular deformable SLAM problem by integrating a Dynamic Deformation Graph (DDG), an evolved form of ED models, that allows us to model general geometries even with discontinuities, in conjunction with visco-elastic deformations, to model better the deformations occurring in real medical scenarios.

\section{NR-SLAM deformation model}
This section is devoted to the deformation model used in NR-SLAM, which is one of the main insights of the system as it is the central piece of all components in NR-SLAM (Fig. \ref{fig::nrslam}): the deformable tracking uses it to compute the current camera pose and the environment deformations in all images, the deformable mapping employs it to extend and refine the map, that keeps track of which points deform together in a graph data structure.

\subsection{Visco-Elastic deformation Model}
Modeling general deformations is a complex problem as it directly depends on the problem domain to be solved. Too general models present observability problems while specific ones can loose capabilities when the observed deformations deviate from the assumptions made. 

For those reasons, we design our deformation model to be general enough to represent different deformations and topologies and, at the same time, be specific enough to our application domain in order to get accurate results. Our deformation model takes into account that in medical scenarios deformations tend to happen slowly and are locally similar in direction and magnitude. Here we present our Visco-Elastic deformation model, representing a spring and a dumper connecting two surface points. On this way, the deformation model is general enough to represent arbitrary surfaces while constraining enough the problem to yield reasonable reconstructions.

First, we make a Camera-over-Deformation (CoD) assumption, that is we assume that most of the image innovation comes from camera motion rather than from deformations of the environment. This assumption, a relaxation of the rigidity assumption of conventional SLAM systems, is usually true when working with temporally close images as in a video sequence from an endoscope. This allows our system to have a reasonable initial seed when trying to initialize and extend the map. Note that this assumption is in sharp contrast with most NRSfM methods that assume a static camera.

Second, we base our deformable model in a set of local deformations modeled as 3D displacements. We consider that small local surface areas tend to behave as almost rigid, while in a more global context deformations can be larger. This is a reasonable prior in medical scenarios, where close points that belong to the same tissue suffer small changes in their relative distances (for example, with muscle stretching), while the relative distance between points that are further apart in the same organ or that belong to different organs suffer bigger changes due to body motions or breathing. Again, this is in contrast with previous NRSfM methods that assume isometric or equiareal deformations. 

In all optimizations performed in our system (deformable tracking and Deformable Bundle Adjustment), we encode our deformation model as a set of two regularizers added to the optimization cost function. First, we use an \textit{Elastic regularizer} that penalizes changes in the distance between two locally close surface points $i$ and $j$ formulated as:

\begin{equation}\label{eq::pos_reg}
    E^t_{ij,elas} = k \frac{(d^t_{ij} - d^0_{ij})^2}{d^0_{ij}}
\end{equation}
where $k$ is a global hyperparameter that controls how strong is the regularizer and $d^t_{ij} = \| \mathbf{x}^t_i - \mathbf{x}^{t}_j \|_2$ is the Euclidean distance between points $i$ and $j$ at time $t$. 

We also add a \textit{Viscous regularizer} that penalizes locally close points that move with different velocities:

\begin{equation}\label{eq::spa_reg}
    E^t_{ij,visc} = b^t_{ij} \| \pmb\delta^t_i - \pmb\delta^t_j \|^2
\end{equation}
where $\pmb\delta^t_i = \mathbf{x}^t_i - \mathbf{x}^{t-1}_i$ is the motion suffered by point $i$ at time $t$ and $b^t_{ij}$ is a pairwise weight between points $i$ and $j$ that models the cross influence between this pair of points computed as:

\begin{equation}\label{eq::weight}
    b^t_{ij} = \exp\left({\frac{-d_{ij,max}^2}{2 \sigma^2}}\right)
\end{equation}
where $d_{ij,max}^2$ is the maximum distance observed between map points $i$ and $j$ and $\sigma$ is a radial basis that controls the influence between points with respect to their relative distance. This weight is directly inspired from the one used in DynamicFusion \cite{newcombe2015dynamicfusion} to regularize the deformation nodes and later used again in \cite{rodriguez2022tracking} to regularize the different points tracked.

\begin{figure*}[t]
\centering
\begin{subfigure}{0.60\columnwidth}
  \centering
  \includegraphics[width=1\columnwidth]{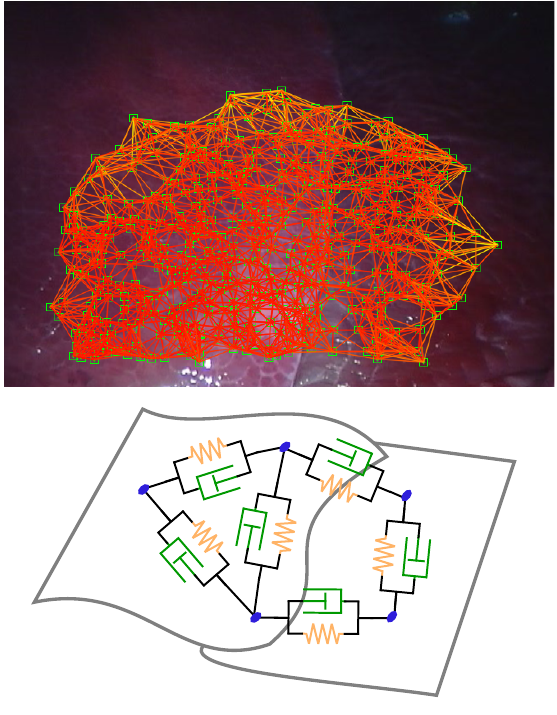}
  \caption{Initial DDG.}
  \label{fig::ddg_repose}
\end{subfigure}
\begin{subfigure}{0.60\columnwidth}
  \centering
  \includegraphics[width=1\columnwidth]{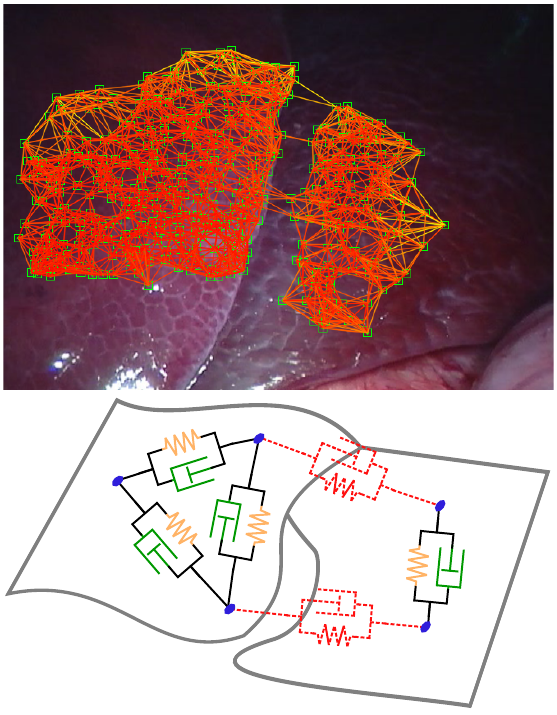}
  \caption{Splitting stage.}
  \label{fig::ddg_split}
\end{subfigure}
\begin{subfigure}{0.60\columnwidth}
  \centering
  \includegraphics[width=1\columnwidth]{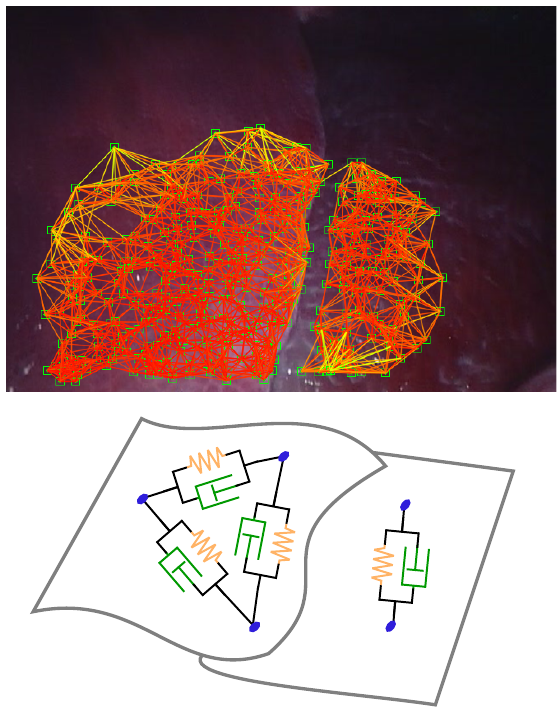}
  \caption{DDG after update.}
  \label{fig::ddg_updated}
\end{subfigure}
\caption{Visual representation of our DDG in a medical sequence where 2 surfaces move independently. At the start, the DDG is built by creating point connections according to their proximity in 3D (Fig. \ref{fig::ddg_repose}). After some deformations, both surfaces move away up to a point in which some connections break according to the criterion presented in section \ref{eq::ddg_criteria} (Fig. \ref{fig::ddg_split}). After that, the DDG is updated and points belonging to different surfaces are not regularized together(Fig. \ref{fig::ddg_updated}).}
\label{fig::ddg}
\end{figure*}

Interestingly enough, this model has a direct mechanical interpretation: the Elastic regularizer acts as a spring connecting two surface points where $k$ in equation \ref{eq::pos_reg} is an homologous to the spring constant found in the Hook's law and the viscous regularizer as a damper where the weight $b^t_{ij}$ in equation \ref{eq::spa_reg} is the proportional constant of a physical damper. This provides our formulation with a physical meaning, enhancing its interpretability.

\subsection{Dynamic Deformation Graph}

Our deformation model relates locally close points in order to constrain their deformations as they tend to locally behave in the same manner. However, this is not always true as there can be points that at some instant can be spatially close but belong to different surfaces that deform differently. Regularizing those points together would lead to incorrect deformation estimations. In order to control this, we propose to encode surface point relationships inside a graph structure that we call Dynamic Deformation Graph (DDG). Basically, DDG represents points as the graph nodes and the relationships between them as the graph edges, which carry the information of which points should be regularized together.

The DDG is first initialized according to the relative 3D distance between points but is periodically updated as the system observes the deformations of the scene (Fig. \ref{fig::ddg}). This leads to a model that is able to understand the dynamics of different surfaces and integrates them correctly into our optimization backbone. The criteria used to prune a connection in the DDG is based in a simple geometric test: if the distance between two connected points grows beyond some streching threshold, points are considered to not behave in a similar fashion and therefor are disconnected in the DDG. Mathematically, this criteria is expressed as:

\begin{dmath}
    \text{prune edge } i,j \text{ in the DDG if } \\ \frac{d_{i,j,max} - d_{i,j,min}}{d_{i,j,min}} > th_{streching}
\end{dmath}\label{eq::ddg_criteria}

Following with the mechanical interpretation of our deformation model, this criteria can be interpreted as the spring and damper joining two points breaking if the points get too far apart.

\section{Deformable Tracking}
In this section we present the tracking module of NR-SLAM that integrates our deformation model embedded in the DDG into a pipeline that estimates in real time the camera pose and the deformations of the observed scene for each incoming frame.

\subsection{Short term data association}
Accurate inter-frame data association is crucial in order to obtain good accuracy and robustness, but medical images suffer from weak texture, local illumination changes and strong reflections. Direct methods like \cite{engel2017direct} have shown good performance in low texture environments getting point associations as a byproduct of the tracking. However, this is done by imposing a global rigid transformation to all points as the environment is assumed stationary, and by assuming illumination constancy or just a global illumination change. Both assumptions are far from true in endoscopy due to deformations and moving illumination. 

Instead, we use a semi-direct approach that performs photometric data association with Shi-Tomasi features \cite{shi1994good} using the modified multi-scale Lucas-Kanade algorithm proposed in \cite{gomez2021sd}:

\begin{equation}\label{eq::klt_enhan_residual}
\argmin_{\mathbf{d}_i,\alpha_i,\beta_i} \;\; \sum_{\mathbf{v} \in P(\mathbf{u}_i)}(I^{0}(\mathbf{v}) - \alpha_i I^t\left(\mathbf{v} + \mathbf{d}_i) - \beta_i\right)^2 
\end{equation}
where $P(\mathbf{u}_i)$ is a small pixel patch centered at the keypoint $\mathbf{u}_i$; $I^0$ is the first frame, where the points are intialized, and $I^t$ is the current frame at time $t$. These patches are updated every five images to account for big scale changes or rotations. Also a local illumination invariance is achieved by computing local gain $\alpha_i$ and bias $\beta_i$ terms for each point. This algorithm has been proven to achieve excellent results when tracking image features in short time steps even in the presence of deformations or local illumination changes \cite{gomez2021sd}\cite{rodriguez2022tracking}. The key of its performance lies in using no global model: each point can move freely with respect the others. 

In order to remove outlier tracks, we compute the \textit{Structural Similarity Index} (SSIM) \cite{wang2004image} between the reference $x$ and tracked $y$ pixel patches:

\begin{equation}\label{eq::SSIM}
    SSIM(x,y) = \frac{(2\mu_x\mu_y+C_1)(2\sigma_{xy}+C_2)}{(\mu^2_x + \mu^2_y + C_1)(\sigma^2_x+\sigma^2_y+C_2)}
\end{equation} where $\mu_x$ and $\sigma_x$ are the mean and covariance of the pixel patch, $\sigma_{xy}$ is the crossed covariance between both patches and $C_1$ and $C_2$ are constant values to avoid inestability when means and covariances approaches to zero. This has been proven to be a good similarity metric for small pixel windows as it combines in a same metric a luminance, contrast and structure comparison.

\subsection{Camera pose and map deformation tracking}

The goal of the deformation tracking task is to jointly estimate the current camera pose $\mathbf{T}_{C^tW}$ and map point deformations $\pmb\delta_i^t$ for the current input frame $I^t$. With that in mind, we combine our deformation model with a reprojection error obtained from the matches provided by our data association algorithm. The idea behind this is to confine possible solutions in a set that is consistent with what the camera is actually seeing. Therefor, we build our cost function with a reprojection term and a deformation term:


\begin{equation}\label{eq::cost_function}
    \mathcal{E}^t = \sum_{i \in \mathcal{P}} (E^t_{i,rep} + E^t_{i,def})
\end{equation}
where $\mathcal{P}$ represents the set of map points tracked by our data association algorithm in the current frame. The reprojection error term can be expressed as:

\begin{equation}\label{eq::rep_error}
E^t_{i,rep} = \rho \big(\| \mathbf{u}_i^t-\mathbf{\hat{u}}_i^t \|^2_{\Sigma_{rep}} \big)
\end{equation}
where $\rho$ is the Hubber robust cost, $\Sigma_{rep}$ is the uncertainty of the matched image features and $\mathbf{u}_i^t$ and $\mathbf{\hat{u}}_i^t$ are respectively the match of feature $i$ in the current image $I^t$ and its projection given by:

\begin{equation}\label{eq::reprojection}
\mathbf{\hat{u}}_i^t = \Pi(\mathbf{T}_{C^tC^0} ({\mathbf{x}_i^{t-1}+\pmb\delta}_i^t))
\end{equation}

The accuracy of indirect methods is limited by the feature detector resolution (typically no less than 1 pixel). 
However, matches obtained with semi-direct methods provide subpixel accuracy boosting in this way the accuracy of our reprojection term while keeping its nice convergence basin.

The deformation term in eq. \ref{eq::cost_function} encodes our deformation model. More precisely, points in $\mathcal{P}$ are regularized together according to our DDG following the equation below:

\begin{equation}\label{eq::def_error}
    E^t_{i,def} = \sum_{j \in \mathcal{N}_{DDG}(i)} E^t_{ij,elas} + E^t_{ij, visc}
\end{equation}
where $\mathcal{N}_{DDG}(i)$ are the neighbours of point $i$ in the DDG. Goal function in eq. \ref{eq::cost_function} is minimized at frame rate using Weighted Non Linear Squares algorithms in order to find the optimal camera pose and deformations:

\begin{equation}\label{eq::tracking_cost}
    \mathbf{T}_{C^tW}, 
    {\pmb\delta}_i^t = \underset{ \mathbf{T}_{C^tW}, 
    {\pmb\delta}_i^t}{\argmin} \;\; {\mathcal{E}^t}
\end{equation}

As this is a highly non-linear problem, it is really important to choose a good first guess for the solution ti ease the solver convergence. For that, in order to assert smooth camera trajectories, the seed for $\mathbf{T}_{C^tC^0}$ is computed as follows:

\begin{itemize}
    \item First, a coarse estimation $\hat{\mathbf{T}}_{C^tC^0}$ is computed using a constant velocity model. 
    \item Second, using the short term data association and $\hat{\mathbf{T}}_{C^tC^0}$, we run a rigid pose-only optimization to refine the seed.
\end{itemize}

\begin{figure}[t]
\centering
\begin{subfigure}{0.45\columnwidth}
  \centering
  \includegraphics[width=1\columnwidth]{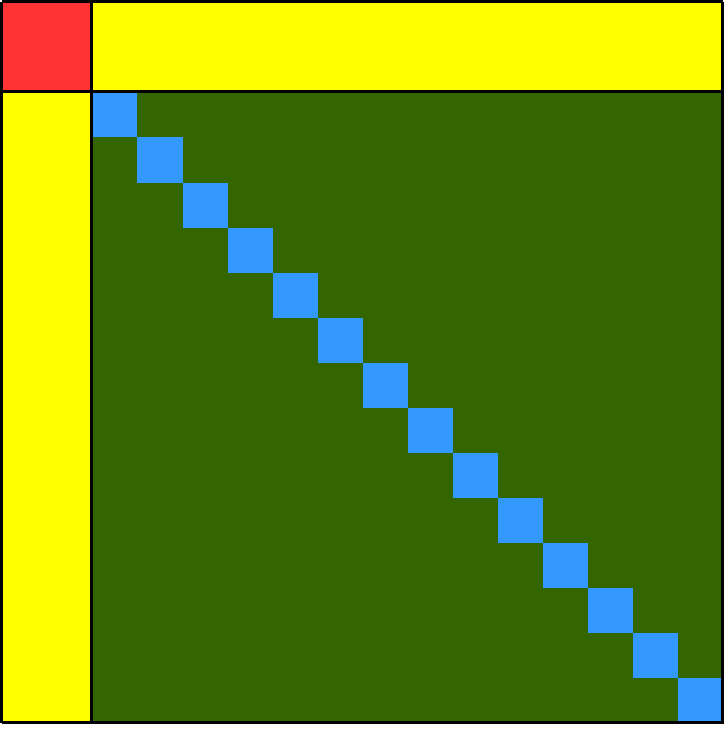}
  \caption{Fully connected\\($D = \infty$)}
  \label{fig::dt_dense}
\end{subfigure}
\begin{subfigure}{0.45\columnwidth}
  \centering
  \includegraphics[width=1\columnwidth]{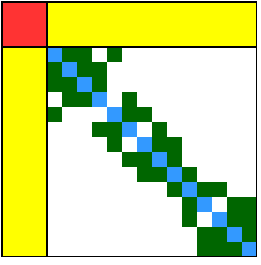}
  \caption{3 best connections\\($D = 3$)}
  \label{fig::dt_sparse}
\end{subfigure}
\caption{Hessian of camera and deformation tracking with different values for the maximum degree $D$ in the DDG. The visco-elastic regularizer densifies the point Hessian.}
\label{fig::dt}
\end{figure}

\begin{figure*}[h]
\centering
\begin{subfigure}{0.9\columnwidth}
  \centering
  \includegraphics[width=1.15\columnwidth]{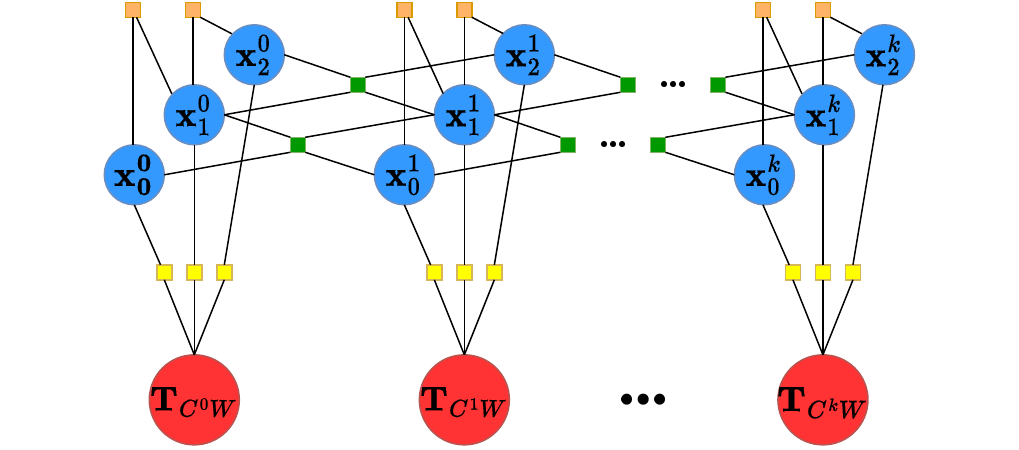}
\end{subfigure}
\begin{subfigure}{0.80\columnwidth}
  \centering
  \includegraphics[width=0.6\columnwidth]{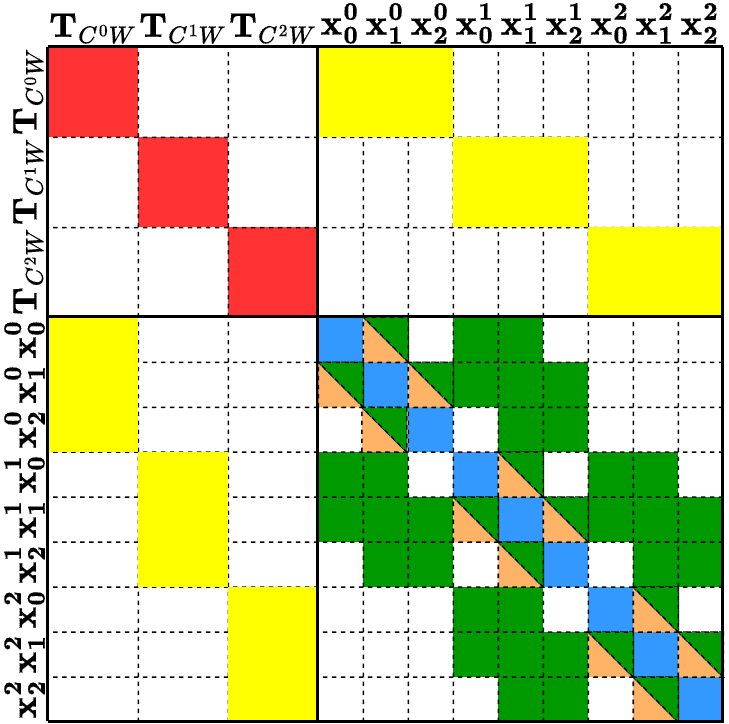}
\end{subfigure}
\begin{subfigure}{0.30\columnwidth}
  \centering
  \includegraphics[width=1\columnwidth]{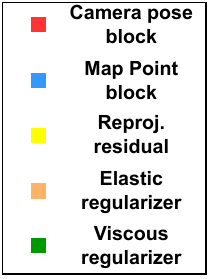}
\end{subfigure}
\caption{Factor graph and Hessian of our proposed Deformable Bundle Adjustment (DBA).}
\label{fig:factor_graphs}
\end{figure*}

The optimization problem stated in eq. (\ref{eq::tracking_cost}) is fairly expensive due to several factors. First, we are estimating $6 + 3|\mathcal{P}|$ degrees of freedom instead of just 6 like in the classical rigid pose estimation problem. Secondly, the introduction of constraints in the form of regularizers between map points (eq. \ref{eq::def_error}) induces off-diagonal blocks in the optimization Hessian, effectively densifying it and increasing the computational cost of the optimization (fig \ref{fig::dt_dense}). In an effort to keep a balance between accurate estimations and computational cost, we decide to set an upper limit $D$ to the degree of nodes in the DDG, which limits the number of regularizers per map point to $D$, sparsifying the optimization Hessian (Fig. \ref{fig::dt_sparse}). We select which points have to be regularized together by selecting the best connections in the DDG according to their weight $b_{ij}^t$ (eq. \ref{eq::weight}).

\subsection{Mid Term Data Association}
Once the current camera pose and map deformation are estimated for the current frame, our algorithm tries to perform a mid-term data association step \cite{campos2021orb} to refine the initial solution by trying to reuse as much data as possible. In other words, map points that have not been used in the first estimation, either because they were lost by our data association algorithm or marked as outliers in previous estimations, are searched in the current image using a guided matching scheme: considering the current geometry, we project into the image candidate map points to be reused. These projections are used as seeds for a Lucas-Kanade tracker that tries to match them in the current image. Those map points successfully matched are integrated into the next tracking estimation, contributing to reducing the overall error.

The key here is the 3D positions used as seeds for the candidate points. There are two cases:

\begin{itemize}
    \item Points updated thanks to the DDG: map points that are neighbours of the currently tracked features in the DDG are updated thanks to the visco-elastic regularizers. 
    \item Other in-frustum map points: for the rest of the points, their last known position is used if it falls within the frustum of the current camera.
\end{itemize}


Having a map available allows us to compute these mid term data associations. This plays a crucial role in our system as it greatly helps to improve the reconstruction results by reusing as much information as possible from the environment already explored.

\section{Deformable Mapping}
Mapping while exploring deformable environments is a crucial but hard task whose goals are to to create, extend and refine the map used by the deformable tracking part. In this section, we describe the basic functions performed by our proposed deformable mapping algorithm.

\subsection{Deformable Bundle Adjustment}

Inspired by state of the art SLAM algorithms, we perform a refinement of the map each time a keyframe is inserted by running a Bundle Adjustment. This operation has been shown to be too expensive to be performed at frame rate,
 and it is usually applied to a selected set of keyframes $\mathcal{K}$ and map points $\mathcal{P}$ to reduce the computational burden. $\mathcal{K}$ and $\mathcal{P}$ in rigid systems are selected using a covisibility criteria around the latest keyframe to maximize the information gained during the optimization \cite{mur2015orb}. However in this work, due to the nature of continuous deformations, $\mathcal{K}$ is built from the latest $N_k$ keyframes in a sliding window fashion as we require them to be consecutive to correctly represent our deformation model.

This optimization, which we coin as Deformable Bundle Adjustment, aims to minimize the following objective function:

\begin{equation}\label{eq::dba}
    \mathcal{E}_{DBA} = \sum_{t \in \mathcal{K}} \sum_{i \in \mathcal{P}^t} (E^t_{i,rep} + E^t_{i,def})
\end{equation}
where $E^t_{i,rep}$ and $E^t_{i,def}$ are the same error terms as in eq. (\ref{eq::cost_function}) and $\mathcal{P}^t$ is the set of points observed from keyframe $t$. This objective function is noticeably more expensive than a regular rigid Bundle Adjustment. First, the number of variables to estimate increases from $|\mathcal{K}| + |\mathcal{P}|$ to $|\mathcal{K}| \cdot |\mathcal{P}|$. Second, the cost of classical Bundle Adjustment is lineal in the number of points as one can apply the Schur trick (marginalizing map points using Schur complement) as the point-point section of the Hessian is block diagonal. In the deformable case, the introduction of regularizers between points adds off-diagonal blocks in the point-point section of the Hessian (Fig. \ref{fig:factor_graphs}). This densification destroys the benefit of the Schur trick, increasing the computational cost of the DBA with respect to its rigid counterpart.

\subsection{Monocular Map Initialization}
In order to track a camera and the deformations, first the mapping needs to create a map. In settings with stereo or RGB-D sensors, a map can be easily initialized from a single view. In rigid monocular SLAM this can be performed with Structure-from-Motion (SfM) algorithms to get a first map estimation up to an unknown scale factor.

However, in a deforming environment with a set of freely moving points, SfM has infinitely many solutions.  Authors in the literature have proposed different Non-Rigid Structure from Motion algorithms (NRSfM), that need to assert a set of strong conditions to constraint the solution like orthographic cameras, stationary cameras, isometric or area-preserving deformations, or close-to-planar surfaces. In most practical deformable SLAM scenarios, and in particular, in medical SLAM, these assumptions are invalid.

Instead, we propose a new map initialization method for deformable environments based on the CoD assumption: most image innovation comes from the camera motion rather than from deformations. With that in mind, our problem statement is to compute the relative camera position between two temporally close images $\mathbf{T}_{C^t,C^0}$ and the 3D positions of the observed map points in each of the images $\mathbf{x}_i^{0}$, $\mathbf{x}_i^{t}$.

Our proposed approach is a two step algorithm. First, using the CoD assumption, we compute an initial estimation of the map assuming using SfM techniques as if the scene was rigid. For that, we track a set of keypoints in the images and estimate from their projection rays an Essential Matrix at low parallax which is decomposed to obtain the relative pose of both cameras $\mathbf{T}_{C^t,C^0}$ that is used to perform a rigid triangulation of the matched keypoints. Second, we perform a Deformable Bundle Adjustment to refine the estimated geometry to adapt it to the possible deformations not considered in the first step. As our Deformable Bundle Adjustment makes use of the DDG we need to initialize it in a naive way using no geometrical information as no previous map is available. To cope with that, we cluster \cite{ester1996density} feature tracks according to their optical flow shape to build a preliminary DDG connecting points that were clustered together. 

The key to the performance of our proposed algorithm lies in the idea that under small deformations, rigid algorithms can get a reasonable first estimation of the geometry. For that, it is crucial to initialize the map as soon as possible to reduce the magnitude of any deformations observed. This implicitly means working under small parallax which can easily worsen the initial rigid estimation. We address this by combining the subpixel accuracy of our semi-direct data association algorithm with the weighted mid point triangulation algorithm proposed in \cite{lee2019triangulation} as it provides good 3D estimations under fairly small parallax ($\backsim$0.3 degrees). This triangulation method also has an interesting property: it is rotational invariant and is easily applied to any type of camera model either pinhole or fish-eye, as it uses projection rays, making it suitable for a wide range of applications.

Once a map has been successfully created, we need to initialize our DDG with the estimated geometry. This also requires setting a specific $\sigma$ to compute the weights (eq. \ref{eq::weight}) between points. We make this process fully automatic by scaling the map to have a predefined mean depth (in colonoscopies we use 3cm) and $\sigma$ is set as the depth standard deviation of the scaled map. 

\subsection{Map Point Triangulation }\label{subsec::triangulation}
As the camera explores new regions of the scene, its pose is estimated by the deformable tracking method and it is necessary to add new points in the map to cover the new regions. In rigid environments, points can be triangulated from two views with enough parallax. 

In deformable environments, there can be small time windows where the environment remains stationary enough for rigid triangulation to give consistent depth estimations. However, in strong deformation periods, deformation and camera motion estimations can be coupled together and rigid triangulation would be inaccurate.
For that reason, we run in parallel two triangulation methods, one rigid and one deformable, and apply a model selection strategy to select the best one. 

Both procedures start by tracking during a set of frames $\mathcal{F}$ a set of candidate features to be triangulated $\mathcal{C}$. Note that we perform triangulation at frame rate looking for a compromise between having enough parallax but not too much deformation that could degrade the results.

\subsubsection{Rigid Triangulation}
For each candidate track in $\mathcal{C}$ we take the first $\mathcal{F}_f$ and last frame $\mathcal{F}_l$ where it was observed and check that the mean map point deformation in all intermediate images is less than a threshold. If this rigidity check is satisfied, we apply the weighted mid point  algorithm \cite{lee2019triangulation} with the first and last frame to triangulate the 3D point, for which we also check parallax and reprojection errors.


\subsubsection{Deformable Triangulation}
We propose a novel approach to triangulate new map points observed during deformable periods. This new algorithm relies on the fact that close points should deform together and so, have a similar 3D flow in the same period of time. With that in mind, for feature in $\mathcal{C}$ we try to adjust a 3D trajectory $\mathbf{X}_i = [\mathbf{x}_i^t, \mathbf{x}_i^{t+1}, ..., \mathbf{x}_i^{t+n}]$ so that:

\begin{itemize}
    \item it has a low reprojection error in the frames of $\mathcal{F}$ in which the feature has been matched, and
    \item the 3D trajectory $\mathbf{X}_i$ is similar to those of some predefined neighbour map points $\mathcal{N}_i$. Since up to this point there is no geometry available, neighbours in $\mathcal{N}$ are selected by feature proximity in the last image. 
\end{itemize}

We then perform the estimation of the 3D trajectory of the new landmark by solving the following minimization problem that models our assumptions:

\begin{equation} 
    \mathbf{X}_i = \underset{\mathbf{X}_i}{\argmin} \ {\mathcal{E}_{i,tria}}
\end{equation}
where:
\begin{equation} \label{eq::triangulation}
\begin{split}
    \mathcal{E}_{i,tria} = \sum_{t \in \mathcal{F}} E_{i,rep}^t + E_{i,visc}^t \\ 
    E^t_{i,visc} = \sum_{j \in \mathcal{N}_i} E_{ij,visc}^t
\end{split}
\end{equation}

\begin{table*}[]
\centering
\caption{Comparison of monocular SLAM methods in simulated colonoscopies with different deformations.}\label{tb::simulation}
\begin{tabular}{|c|r|r|r|r|r|r|}
\cline{1-3} \cline{5-7}
Sequence                     & $A$ (mm)              & $\omega$ (rad/s)     &                             & ORB-SLAM3     & SD-DefSLAM & NR-SLAM       \\ \hline \hline
\multirow{2}{*}{simulated-0} & \multirow{2}{*}{0.0}  & \multirow{2}{*}{0.0} & RMSE (mm)                   & \textbf{3.52} & 11.67      & 3.87          \\ \cline{4-7} 
                             &                       &                      & \multicolumn{1}{l|}{\# Fr.} & 84            & 73         & 84            \\ \hline \hline
\multirow{2}{*}{simulated-1} & \multirow{2}{*}{2.5}  & \multirow{2}{*}{2.5} & RMSE (mm)                   & 4.30          & 11.96      & \textbf{2.26} \\ \cline{4-7} 
                             &                       &                      & \multicolumn{1}{l|}{\# Fr.} & 84            & 27         & 84            \\ \hline \hline
\multirow{2}{*}{simulated-2} & \multirow{2}{*}{2.5}  & \multirow{2}{*}{5.0} & RMSE (mm)                   & 5.63          & 11.49      & \textbf{2.72} \\ \cline{4-7} 
                             &                       &                      & \multicolumn{1}{l|}{\# Fr.} & 84            & 78         & 84            \\ \hline \hline
\multirow{2}{*}{simulated-3} & \multirow{2}{*}{5.0}  & \multirow{2}{*}{2.5} & RMSE (mm)                   & 5.09          & 12.67      & \textbf{2.48} \\ \cline{4-7} 
                             &                       &                      & \multicolumn{1}{l|}{\# Fr.} & 84            & 27         & 84            \\ \hline \hline
\multirow{2}{*}{simulated-4} & \multirow{2}{*}{5.0}  & \multirow{2}{*}{5.0} & RMSE (mm)                   & 8.11          & 11.33      & \textbf{3.95} \\ \cline{4-7} 
                             &                       &                      & \multicolumn{1}{l|}{\# Fr.} & 84            & 29         & 84            \\ \hline \hline
\multirow{2}{*}{simulated-5} & \multirow{2}{*}{10.0} & \multirow{2}{*}{2.5} & RMSE (mm)                   & 9.71          & 11.71      & \textbf{3.80} \\ \cline{4-7} 
                             &                       &                      & \multicolumn{1}{l|}{\# Fr.} & 84            & 84         & 84            \\ \hline \hline
\multirow{2}{*}{simulated-6} & \multirow{2}{*}{10.0} & \multirow{2}{*}{5.0} & RMSE (mm)                   & 15.62         & 38.27      & \textbf{5.82} \\ \cline{4-7} 
                             &                       &                      & \multicolumn{1}{l|}{\# Fr.} & 56            & 71         & 84            \\ \hline 
\end{tabular}
\end{table*}

Terms $E_{i,rep}^t$ and $E_{ij,visc}^t$  are the reprojection error and the Viscous regularizer presented in eqs. \ref{eq::rep_error} and \ref{eq::spa_reg} respectively. As the problem  is non-linear, the seed used for the 3D trajectory $\mathbf{X_i}$ may have a strong impact in the estimation output. For that, assuming that surfaces are locally smooth, we initialize each individual point $\mathbf{x}_i^t$ in $\mathbf{X}_i$ by unprojecting the tracked feature and setting its depths as the average depth of the neighbours $\mathcal{N}_i$. In order to accept a deformable triangulation as good, we check that the reprojection error $E_{i,rep}^t$ in each one of the frames $\mathcal{F}$ is lower than a given threshold. We also verify that the estimated trajectory $\mathbf{X}_i$ respects the viscous regularizer, that is, at least half of the viscous regularizers used $E_{ij,visc}^t$ are below a threshold error.

\subsubsection{Model Selection}
From $\mathcal{C}$ and $\mathcal{F}$, we apply both triangulation algorithms, obtaining a set of successful rigid triangulations $\mathcal{R}$ and a set of successful deformable triangulations $\mathcal{D}$. To decide which method got the best result we apply the following model selection method:

\begin{enumerate}
    \item accept $\mathcal{R}$ if $|\mathcal{R}| > 1.5 \, |\mathcal{D}|$
    \item accept $\mathcal{D}$ if $|\mathcal{D}| > 1.5 \, |\mathcal{R}|$
    \item otherwise no clear winner is found and then, no triangulations are accepted.
\end{enumerate}

Once inserted, the new features will be refined in the next DBA when a new keyframe is inserted in the map.

\section{Experiments}

We evaluate our proposed deformable SLAM system in different datasets to assess its capabilities. Our main focus is medical sequences, for that reason, we make use of the Hamlyn dataset \cite{Mountney2010ThreeDimensionalTD} and the Endomapper dataset \cite{azagra2022endomapper}, two of the main datasets used in medical SLAM research. Specifically, we performed evaluations on:
\begin{enumerate}
    \item \textbf{Endomapper simulation sequences} of colonoscopies that allow us to evaluate the camera pose estimation and the accuracy of reconstruction with different degrees of deformation.
    \item \textbf{Hamlyn real sequences} acquired with a stereo endoscope, that allow us to evaluate  reconstruction accuracy under different deformations and topologies.
    \item \textbf{Endomapper real sequences} of colonoscopies, that allow qualitative performance evaluation on real medical scenarios.
\end{enumerate}

Quantitative comparison with other state of the art SLAM algorithms for deformable scenarios is performed using the Hamlyn dataset. We compare with the reconstruction performance of deformable monocular methods in the literature, namely DefSLAM \cite{lamarca2018camera} and SD-DefSLAM \cite{gomez2021sd}, that are full SLAM methods, and DSDT \cite{lamarca2021direct} and MCPD \cite{rodriguez2022tracking}, that are just deformable tracking methods. The main characteristics of these algorithms can be found in table \ref{tab::def_systems}. In some cases, we also include in the comparison ORB-SLAM3 \cite{campos2021orb} as a rigid monocular SLAM system. We refer the reader to the provided video to have a better visualization of the results.

\subsection{Implementation details}
The method was fully implemented from scratch in C++ and runs entirely on CPU. We use the OpenCV library \cite{bradski2000opencv} for computer vision and image processing tasks, Sophus library \cite{Sophus} for representing SE(3) objects and g2o library \cite{kummerle2011g} to implement non-linear least squares optimizations. An open-source version of NR-SLAM is available for the benefit of the community.

\subsection{Quantitative evaluation in simulated colonoscopies}
For evaluating the performance of deformable SLAM methods, we provide in the Endomapper dataset \cite{azagra2022endomapper} a set of simulated colonoscopies. These sequences were obtained with the VR-Caps simulator \cite{incetan2021vr} using their colon model, which was obtained from a Computerized Tomography of a patient, adding photorealistic textures. The sequences simulate a colonoscope insertion manoeuvre with a forward motion, with synthetic deformations added to the scene. The deformations are modelled via a sine wave propagated along the colon model according to the following formula:
\begin{equation}
    V_y^t = V_y^0 + A \sin(\omega t + V_x^0 + V_y^0 + V_z^0)
\end{equation}
where $V_x^0$, $V_y^0$ and $V_z^0$ are the coordinates of a surface point at rest. The strength of the deformations is controlled by the sine amplitude $A$ and its frequency $\omega$. We evaluate the reconstruction accuracy of NR-SLAM under different combinations of values of $A$ and $\omega$, enabling us to test our system under different deformation conditions, from smaller to more sudden and aggressive deformations. 

Results are presented in table \ref{tb::simulation}. With a static scene (sequence-0), NR-SLAM is slightly worse than ORB-SLAM3. This suggests that the synthetic texture used is very good for ORB features and the deformability assumption, when false, plays against our method. With mild deformations (sequences 1 to 4) the accuracy of NR-SLAM actually increases, doubling that of rigid methods. With stronger deformations, that are unbearable for rigid methods, our method still maintains good accuracy. 

\begin{figure*}
\centering
  \def\svgwidth{2.0\columnwidth}
\begingroup%
  \makeatletter%
  \providecommand\color[2][]{%
    \errmessage{(Inkscape) Color is used for the text in Inkscape, but the package 'color.sty' is not loaded}%
    \renewcommand\color[2][]{}%
  }%
  \providecommand\transparent[1]{%
    \errmessage{(Inkscape) Transparency is used (non-zero) for the text in Inkscape, but the package 'transparent.sty' is not loaded}%
    \renewcommand\transparent[1]{}%
  }%
  \providecommand\rotatebox[2]{#2}%
  \newcommand*\fsize{\dimexpr\f@size pt\relax}%
  \newcommand*\lineheight[1]{\fontsize{\fsize}{#1\fsize}\selectfont}%
  \ifx\svgwidth\undefined%
    \setlength{\unitlength}{1791.67755367bp}%
    \ifx\svgscale\undefined%
      \relax%
    \else%
      \setlength{\unitlength}{\unitlength * \real{\svgscale}}%
    \fi%
  \else%
    \setlength{\unitlength}{\svgwidth}%
  \fi%
  \global\let\svgwidth\undefined%
  \global\let\svgscale\undefined%
  \makeatother%
  \begin{picture}(1,0.52467019)%
    \lineheight{1}%
    \setlength\tabcolsep{0pt}%
    \put(0,0){\includegraphics[width=\unitlength,page=1]{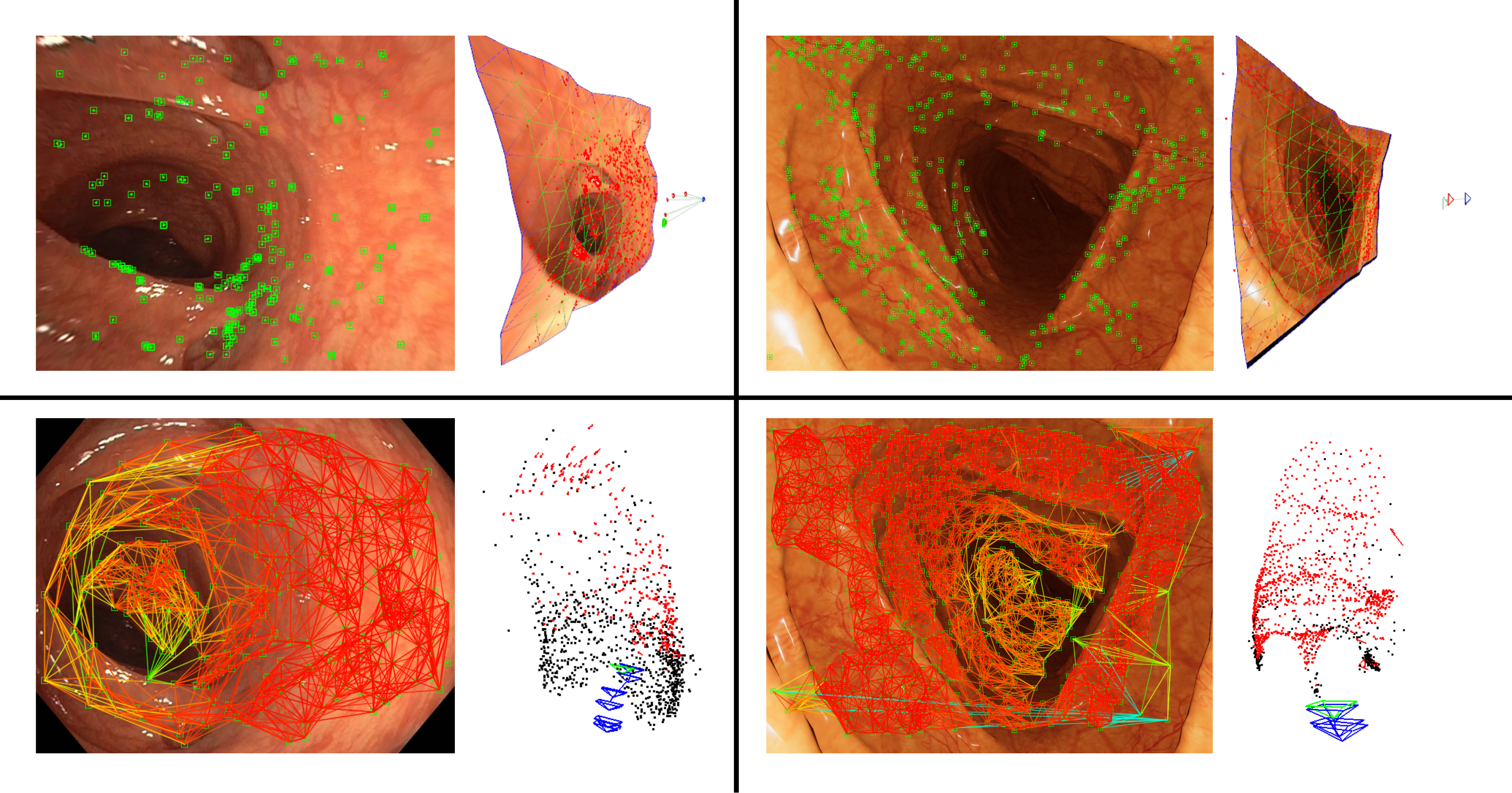}}%
    \put(0.16914546,0.51144768){\color[rgb]{0,0,0}\makebox(0,0)[lt]{\lineheight{1.25}\smash{\begin{tabular}[t]{l}Real Colon Sequence\end{tabular}}}}%
    \put(0.63404796,0.51144737){\color[rgb]{0,0,0}\makebox(0,0)[lt]{\lineheight{1.25}\smash{\begin{tabular}[t]{l}Simulated-3 Colon Sequence\end{tabular}}}}%
    \put(0.01301427,0.33856138){\color[rgb]{0,0,0}\rotatebox{90}{\makebox(0,0)[lt]{\lineheight{1.25}\smash{\begin{tabular}[t]{l}SD-DefSLAM\end{tabular}}}}}%
    \put(0.01301427,0.09277702){\color[rgb]{0,0,0}\rotatebox{90}{\makebox(0,0)[lt]{\lineheight{1.25}\smash{\begin{tabular}[t]{l}NR-SLAM\end{tabular}}}}}%
  \end{picture}%
\endgroup%

  \caption{Comparison between SD-DefSLAM and NR-SLAM in a real and a simulated colonoscopy for Endomapper dataset. While SD-DefSLAM estimates an almost-planar shape, NR-SLAM successfully reconstructs the correct tubular shape.}\label{fig::sd}
\end{figure*}
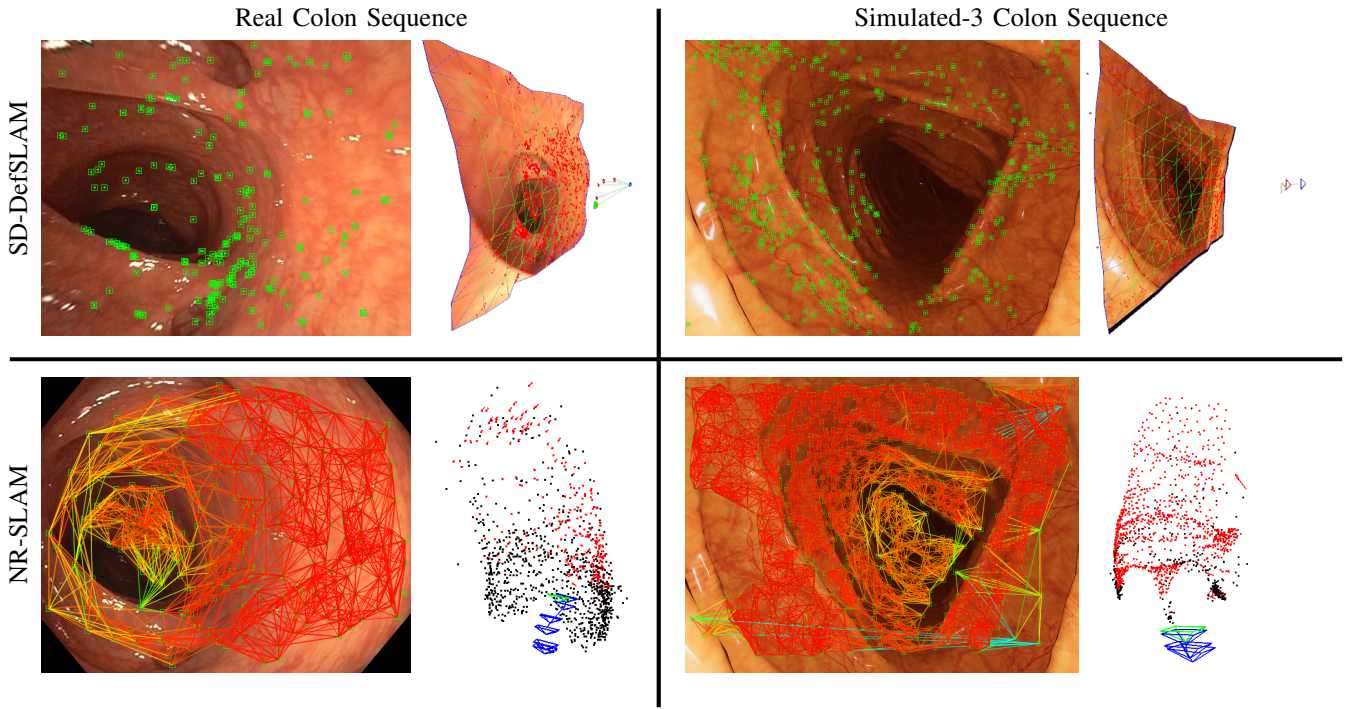

Interestingly enough, SD-DefSLAM, that is the current state of the art in deformable SLAM for medical applications, loses track in most sequences, and obtains poor accuracy, being worst than rigid methods. The reason lies in its  strong assumptions: (1) that the environment is a continuous surface with planar topology and (2) that deformations are isometric. The first one is clearly violated in colonoscopies (see Fig. \ref{fig::sd}) where discontinuities due to haustra and the tubular topology and shape seem to have a huge impact, causing poor accuracy even with a static scene. In comparison, the isometric assumption only seems to have a big impact under strong deformations.

\begin{table*}
\caption{Comparison of monocular tracking methods, with stereo initialization and no mapping.}
\label{tab::stereo_initialization}
\centering
\begin{threeparttable}
\begin{tabular}{ll|r|r|r|r|r|}
\cline{3-7}
\multicolumn{2}{l|}{} & {Rigid map}&
\multicolumn{4}{c|}{Deformable map}\\
\cline{3-7}
\multicolumn{2}{l|}{}                                          & ORB-SLAM3  & SD-DefSLAM  & DSDT & MCPD & NR-SLAM\\ \hline
\multicolumn{1}{|l|}{\multirow{2}{*}{abdominal-sequence-6}}  & RMSE (mm)      & 4.85   & 2.72        & 3.17 & -- & \textbf{2.37} \\ \cline{2-7} 
\multicolumn{1}{|l|}{}                            & \# Fr. & 128   & 286    & 300  & -- & \textbf{1236} \\ \hline \hline
\multicolumn{1}{|l|}{\multirow{2}{*}{exploration-sequence-20}} & RMSE (mm)     & 1.37  & 4.68     & 2.90 & 1.48 & \textbf{1.09} \\ \cline{2-7} 
\multicolumn{1}{|l|}{}                            & \# Fr. & 220    & 252         & 500 & 350 & \textbf{609} \\ \hline \hline
\multicolumn{1}{|l|}{\multirow{2}{*}{liver-sequence-21}} & RMSE  (mm)    & -- & 6.19        & 1.30 & 1.55 & \textbf{0.59} \\ \cline{2-7} 
\multicolumn{1}{|l|}{}                            & \# Fr. & --  & 323          & 300 & 300 & \textbf{376} \\ \hline
\end{tabular}
\end{threeparttable}
\end{table*}

\subsection{Quantitative evaluation in real sequences}

\begin{table*}
\caption{Comparison of full monocular deformable SLAM methods
with initialization, tracking and mapping.}
\label{tab::slam_results}
\centering
\begin{threeparttable}
\begin{tabular}{ll|r|r|r|}
\cline{3-5}
\multicolumn{2}{l|}{}                                          & DefSLAM  & SD-DefSLAM  & NR-SLAM\\ \hline
\multicolumn{1}{|l|}{\multirow{2}{*}{abdominal-sequence-1}}  & RMSE (mm)      & 23.98   & 22.20         & \textbf{12.26}  \\ \cline{2-5} 
\multicolumn{1}{|l|}{}                            & \# Fr. & 858   & 958    &  \textbf{1030} \\ \hline \hline
\multicolumn{1}{|l|}{\multirow{2}{*}{organs-sequence-19}} & RMSE (mm)     & 13.02  & \textbf{6.63}     & 11.09  \\ \cline{2-5} 
\multicolumn{1}{|l|}{}                            & \# Fr. & 1300    & 1300         &  1300 \\ \hline \hline
\multicolumn{1}{|l|}{\multirow{2}{*}{exploration-sequence-20}} & RMSE  (mm)    & 17.02 & 12.56 & \textbf{7.13} \\ \cline{2-5} 
\multicolumn{1}{|l|}{}                            & \# Fr. & 1579  & 1750          & \textbf{1811}  \\ \hline
\end{tabular}
\end{threeparttable}
\end{table*}

We perform a quantitave study of the reconstruction performance of our proposed system and other state of the art algorithms with the Hamlyn dataset \cite{Mountney2010ThreeDimensionalTD}. This dataset is composed of several laparoscopic calibrated stereo sequences that can be used to get depth ground-truth. From all of them, we select the following sequences:

\begin{itemize}
    \item \textbf{abdominal-sequence-1}: a quasi-rigid sequence of an abdominal cavity explored with a smooth camera motion.
    \item \textbf{abdominal-sequence-6}: a quasi-rigid sequence with a close look-up of another abdominal cavity.
    \item \textbf{organs-sequence-19}: a sequence imaging a deforming organ with some tools interfering in the field of view.
    \item \textbf{exploration-sequence-20}: a general exploration sequence of several deforming organs and tissues.
    \item \textbf{liver-sequence-21}: a sequence imaging two lobes of a liver, with significant relative motion due to breathing, as shown in figure \ref{fig::ddg}.
\end{itemize}

We perform two comparisons with this dataset. First, an ablative study of the robustness and accuracy of camera tracking and deformation estimation, given an initial 3D reconstruction obtained with stereo. For this experiment, we compare the tracking part of ORB-SLAM3, DefSLAM, SD-DefSLAM and NR-SLAM, and DSDT and MPCD that are pure tracking algorithms,  in tree sequences. We evaluate the number of successfully tracked frames and the reconstruction RMSE across all those frames, comparing with stereo ground-truth. Results are shown in table \ref{tab::stereo_initialization}. Clearly, the deformable tracking proposed in NR-SLAM obtains superior results in both metrics, setting the state of the art in monocular camera tracking and deformation estimation. This is due to both the quality of the data association and our deformation model. Interestingly enough, our deformable tracking algorithm is similar to the one presented in MCPD, but we get here much better results due to the addition of the DDG and the visco-elastic regularizer. Compared with SD-DefSLAM we get much better accuracy in sequences 20 and 21. Again, the reason lies in the strong assumptions made by SD-DefSLAM. In this case, surfaces are reasonably planar and continuous, but deformations are not isometric, which is well captured by our deformation model.

The second experiment is performed with full SLAM setups,  comparing DefSLAM, SD-DefSLAM and NR-SLAM in three sequences. Again, results in table \ref{tab::slam_results} show that our proposed system outperforms other algorithms, getting significantly more tracked frames and smaller reconstruction errors in sequences 1 and 20. The exception is sequence 19 where SD-DefSLAM is able to get lower errors. However, this was an expected result as this sequence has several surgical tools intrusions, and SD-DefSLAM uses deep learning techniques to segment and mask them. This makes it able to completely ignore surgical tools that worsen significantly the reconstruction accuracy of the other methods.

\subsection{Qualitative evaluation in real sequences}
\begin{figure*}[t]
\centering
\begin{subfigure}{0.65\columnwidth}
  \centering
  \includegraphics[width=1\columnwidth]{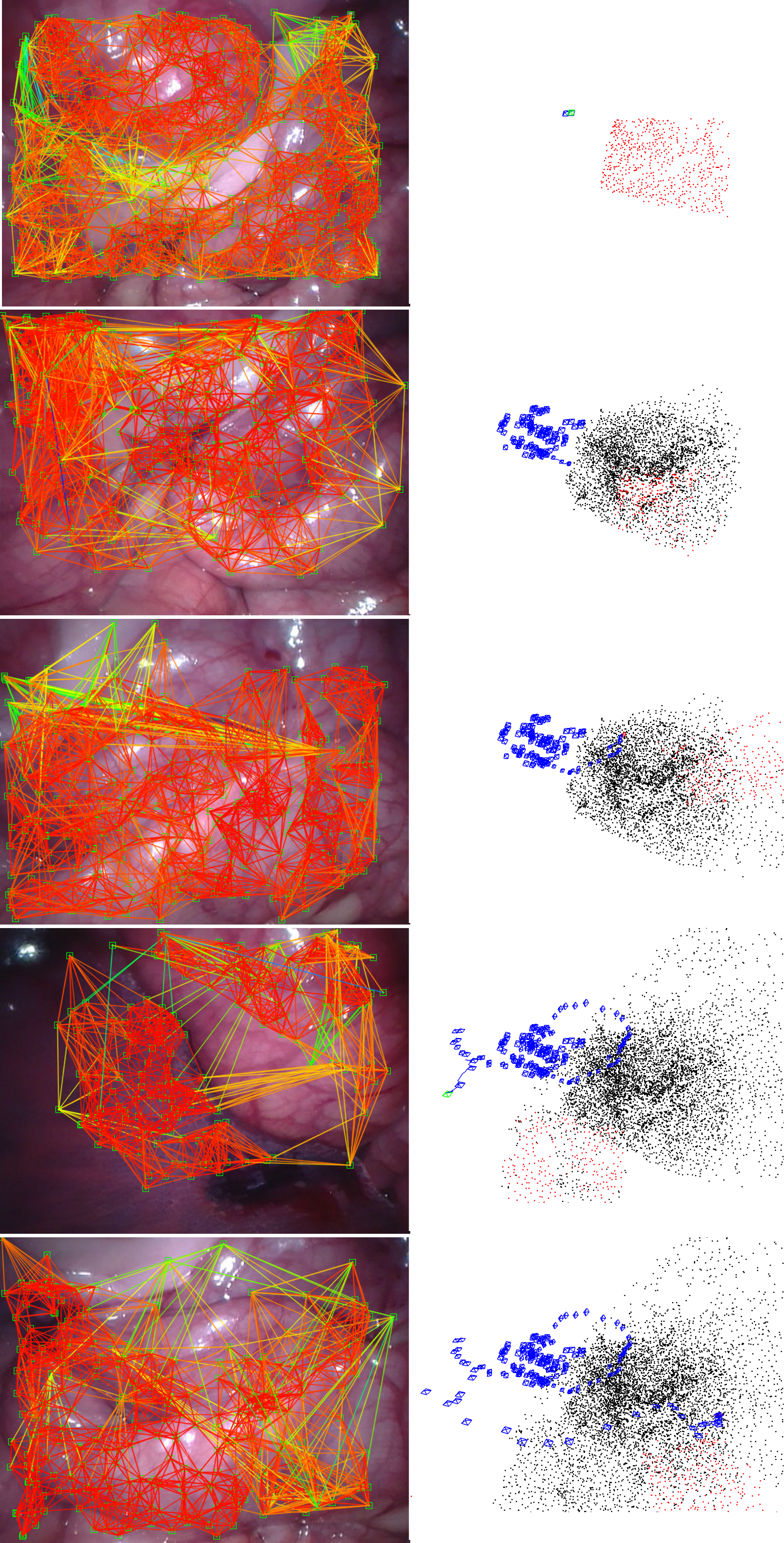}
  \caption{Exploration-sequence-20}
  \label{fig::results_h20}
\end{subfigure}
\begin{subfigure}{0.65\columnwidth}
  \centering
  \includegraphics[width=1\columnwidth]{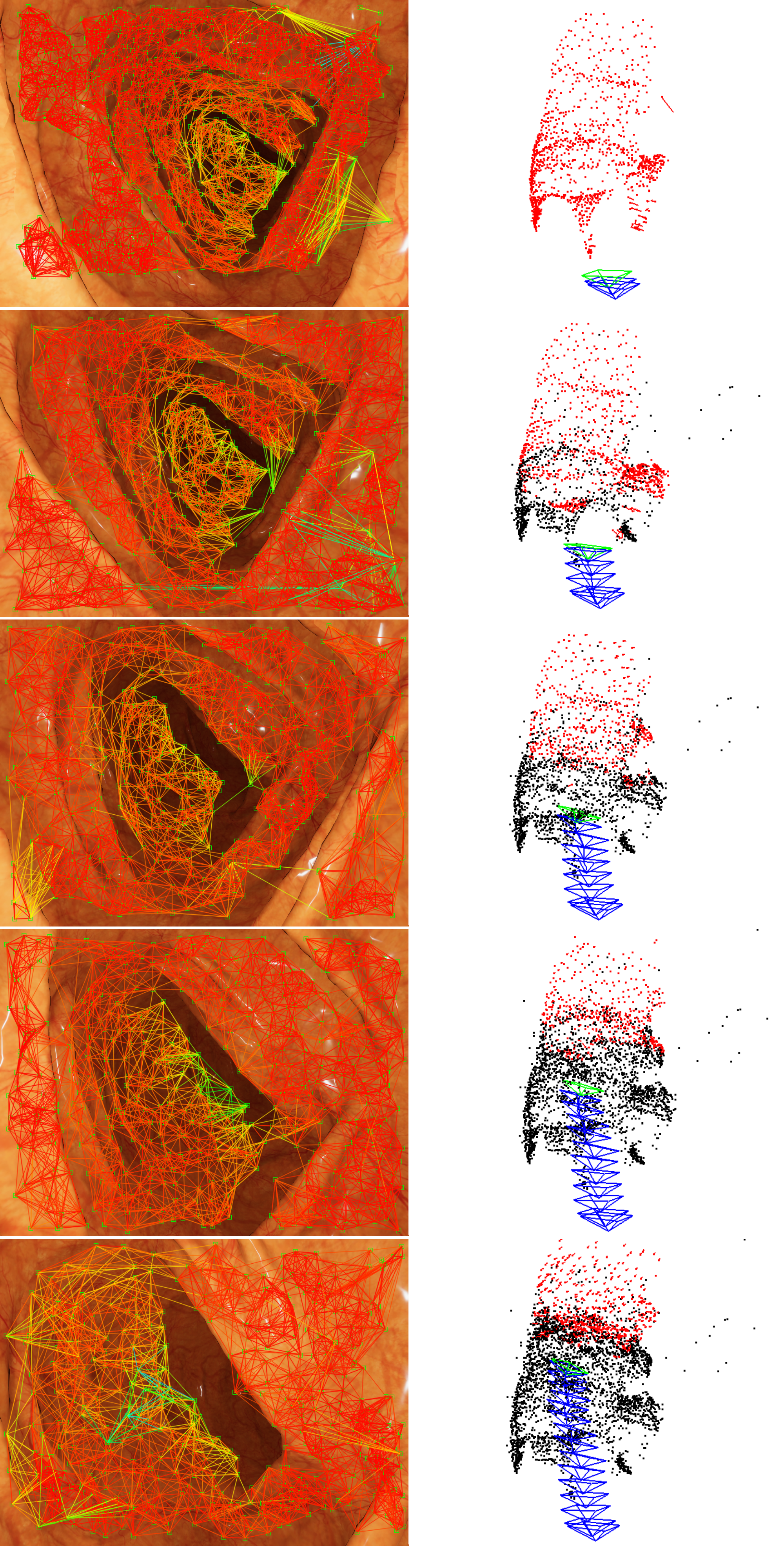}
  \caption{Simulated-3 colonoscopy}
  \label{fig::results_simulation}
\end{subfigure}
\begin{subfigure}{0.65\columnwidth}
  \centering
  \includegraphics[width=1\columnwidth]{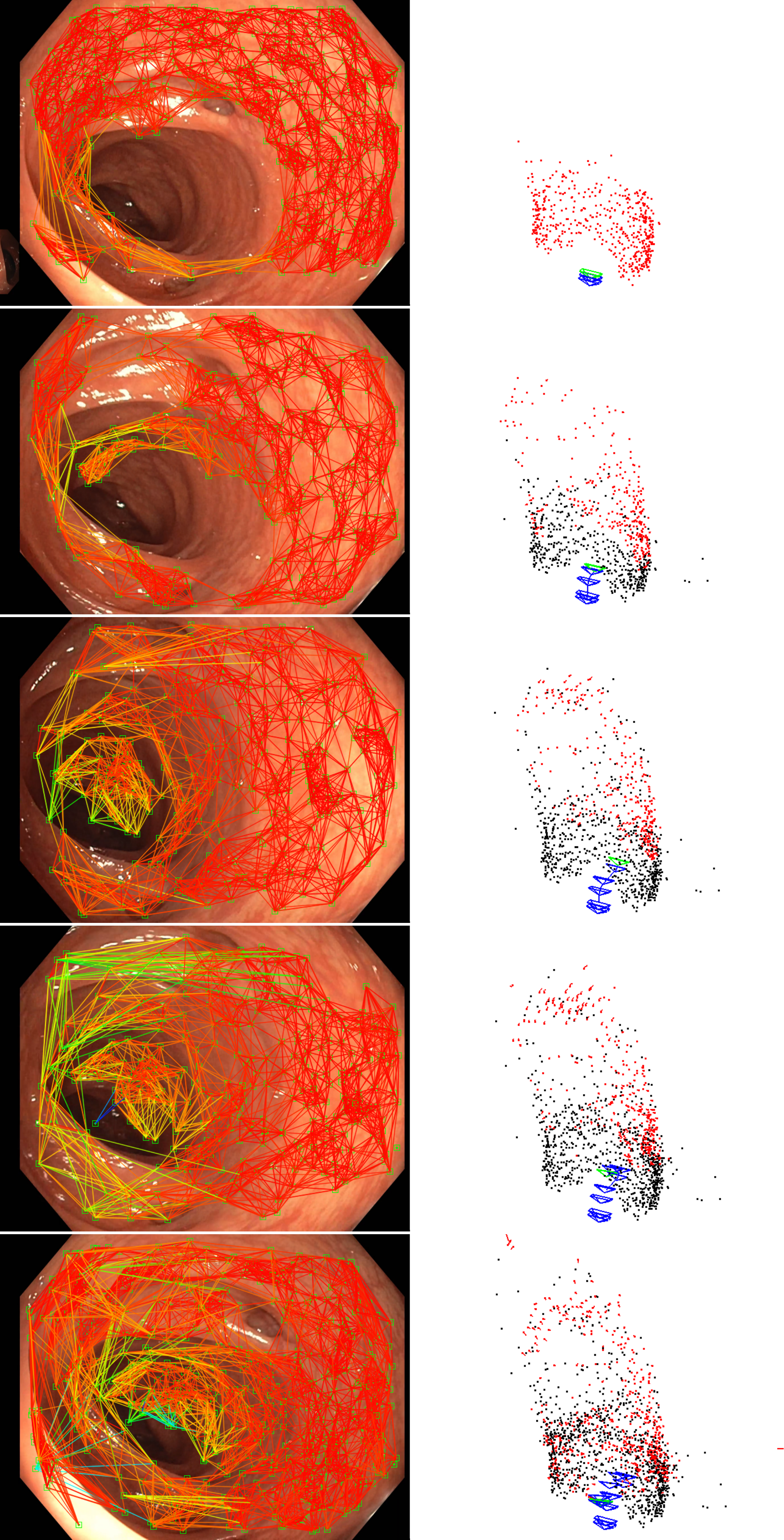}
  \caption{Real colonoscopy}
  \label{fig::results_colon}
\end{subfigure}
\caption{Example reconstructions obtained by NR-SLAM in Hamlyn and two Endomapper colonoscopies.}
\label{fig::results}
\end{figure*}

We finally provide qualitative results in real in-vivo human colonoscopy sequences from the Endomapper dataset. These sequences only provide the RGB images of the procedures with no labels or ground-truth available. Nevertheless, it is an interesting dataset to qualitative test our proposed system as it contains a whole bunch of challenges typically found in medical imagery: tissue deformations, depth discontinuities, little to no texture, varying lighting conditions, specular reflections and fish-eye optics. Figure \ref{fig::results}c shows how NR-SLAM, despite these challenges, is able to initialize and extend a deformable map represented with a DDG, while tracking the camera pose and scene deformation in a real colonoscopy, going beyond the capabilities of previous deformable SLAM systems.

\section{Conclusions}
We have proposed NR-SLAM, the first monocular deformable SLAM system free from strong assumptions like planar topology, almost-planar shape, continuous surface and isometric or equi-areal deformations. It automatically creates, extends and refines a map that is used by our deformable tracking to estimate the camera pose and the map deformation for each frame. Our Visco-Elastic deformation model has a nice physical interpretation and is able to model the generic deformation appearing in medical environments. Our Dynamic Deformation Graph smoothly handles discontinuities from occlusion, like different colon haustra, or discontinuities that appear during mapping, like two organs moving apart. 

Indeed, our deformable tracking method presents a general, robust and accurate solution for the reconstruction of deforming surfaces, achieving outstanding results. However, deformable mapping still has important open issues. General NRSfM is an undetermined problem, that needs priors to be solved. Previous methods use strong priors with a closed mathematical formulation (isometry, orthographic cameras, ...) but our results show that they are problematic for many medical scenarios. We decided to use milder priors with a clear physical meaning in which the environment undergoes smooth deformations, both temporally and spatially, allowing us to cope with a wider range of scenes. 
Our experiments show good performance in deforming and almost planar environments (Fig. \ref{fig::results_h20}) and, for the first time, in environments with depth discontinuities and mild deformations like colonoscopies (Fig. \ref{fig::results_simulation} and \ref{fig::results_colon}). However, while our deformable mapping outperforms the state-of-the-art methods, it still struggles to create and extend a map in sequences with both, depth discontinuities and strong deformations, as in liver-sequence-21 (Fig. \ref{fig::ddg}). This represents the main limitation of our system: in this type of scenarios it needs a given 3D seed to initialize the map or to add new points to it. Probably, this can be addressed in future work using neural networks that estimate depth from monocular images. Also, robustness can be significantly improved by adding surgical tools segmentation and place recognition techniques to recover from occlusions \cite{gomez2021sd}.

\bibliographystyle{IEEEtran}
\bibliography{IEEEabrv,bibl}

\begin{IEEEbiography}[{\includegraphics[width=1in,height=1.25in,clip,keepaspectratio]{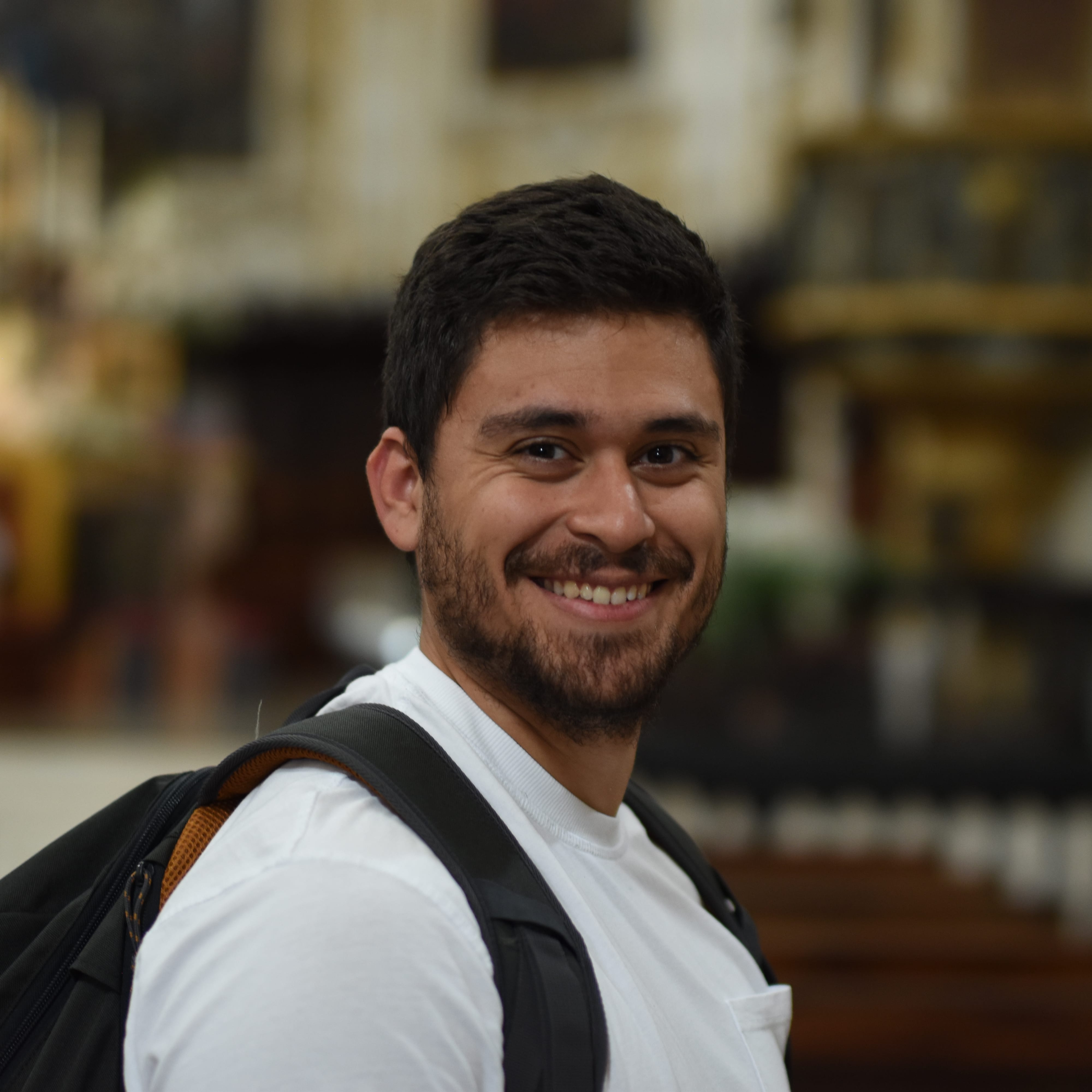}}]{Juan J. Gómez Rodríguez}
received a Bachelor’s Degree in Informatics Engineering (mention in Computing) and Master’s in Biomedical Engineering (mention in Information and Communication Technologies in Biomedical Engineering) from Universidad de Zaragoza, where he is currently working towards the PhD. degree with the I3A Robotics, Perception and Real-Time Group. His research interests are real-time visual SLAM for both rigid and deformable environments. He received an honorable mention to the King-Sun Fu Memorial IEEE Transactions on Robotics Best Paper Award in 2021, for the paper describing ORB-SLAM3.
\end{IEEEbiography}

\begin{IEEEbiography}[{\includegraphics[width=1in,height=1.25in,clip,keepaspectratio]{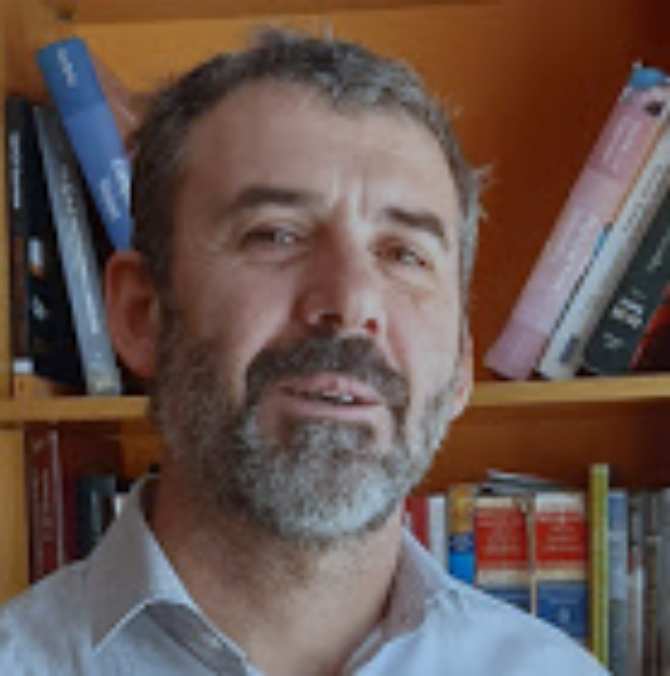}}]{J.M.M. Montiel}
(Arnedo, Spain, 1967) received the M.S. and PhD degrees in electrical engineering from Universidad de Zaragoza, Spain, in 1992 and 1996, respectively. He has been awarded several Spanish MEC grants to fund research with the University of Oxford, U.K., and Imperial College London, U.K.

He is currently a full professor with the Depar- tamento de Informática e Ingeniería  de Sistemas, Universidad de Zaragoza, where he is in charge of perception and computer vision research grants and
courses. His interests include real-time visual SLAM for rigid and non-rigid environments, and the transference of this technology to robotic and non- robotic application domains. He has received several awards, including the 2015 King-Sun Fu Memorial IEEE Transactions on Robotics Best Paper Award and an honorable mention to the same award in 2021. Since 2020 he coordinates the EU FET EndoMapper grant to bring visual SLAM to intracorporeal medical scenes.
\end{IEEEbiography}

\begin{IEEEbiography}[{\includegraphics[width=1in,height=1.25in,clip,keepaspectratio]{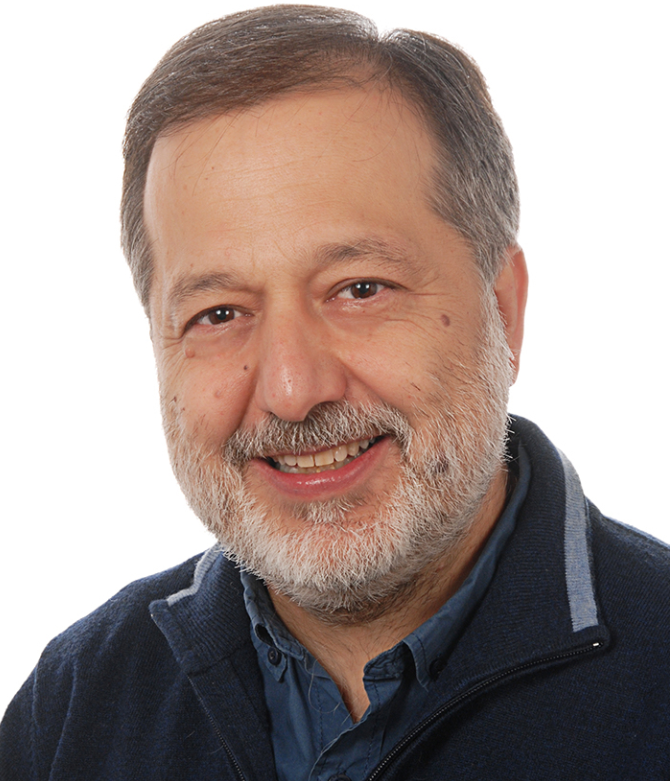}}]{Juan D. Tardós}
(Huesca, Spain, 1961) received the M.S. and Ph.D. degrees in electrical engineering from the University of Zaragoza, Spain, in 1985 and 1991, respectively. He is Full Professor with the Departamento de Informática e Ingeniería de Sistemas, University of Zaragoza, where he is in charge of courses in machine learning and SLAM. His research interests include SLAM, perception and mobile robotics. He received the King-Sun Fu Memorial IEEE Transactions on Robotics Best Paper Award in 2015 for the paper describing the monocular SLAM system ORB-SLAM and an honorable mention to the same award in 2021, for the paper describing ORB-SLAM3.
\end{IEEEbiography}

\end{document}